\pdfoutput=1 
\documentclass{article}

\usepackage[final]{neurips_2023}

\usepackage[utf8]{inputenc} % allow utf-8 input
\usepackage[T1]{fontenc}    % use 8-bit T1 fonts
\usepackage{hyperref}       % hyperlinks
\usepackage{url}            % simple URL typesetting
\usepackage{booktabs}       % professional-quality tables
\usepackage{amsfonts}       % blackboard math symbols
\usepackage{nicefrac}       % compact symbols for 1/2, etc.
\usepackage{microtype}      % microtypography
\usepackage{xcolor}         % colors

\usepackage{algorithm}
\usepackage{algorithmic}

\usepackage[normalem]{ulem}

\title{PaSS: Parallel Speculative Sampling}

\author{%
    Giovanni Monea\thanks{Work done while at Apple} \\
    EPFL \\
    \texttt{giovanni.monea@epfl.ch}
    \And 
    Armand Joulin \\
    Apple \\
    \texttt{ajoulin@apple.com}
    \And
    Edouard Grave \\
    Apple \\
    \texttt{egrave@apple.com}
}

\begin{document}

\maketitle

% MODIFIED ABSTRACT
\begin{abstract}
Scaling the size of language models to tens of billions of parameters has led to impressive performance on a wide range of tasks. At generation, these models are used auto-regressively, requiring a forward pass for each generated token, and thus reading the full set of parameters from memory. This memory access forms the primary bottleneck for generation and it worsens as the model size increases. Moreover, executing a forward pass for multiple tokens in parallel often takes nearly the same time as it does for just one token. These two observations lead to the development of speculative sampling, where a second smaller model is used to draft a few tokens, that are then validated or rejected using a single forward pass of the large model. Unfortunately, this method requires two models that share the same tokenizer and thus limits its adoption. 
As an alternative, we propose to use parallel decoding as a way to draft multiple tokens from a single model with no computational cost, nor the need for a second model. 
Our approach only requires an additional input token that marks the words that will be generated simultaneously.
We show promising performance (up to $30\%$ speed-up) while requiring only as few as $O(d_{emb})$ additional parameters.
\end{abstract}

% ORIGINAL ABSTRACT
% \begin{abstract}
% Scaling the size of language models to tens of billions of parameters have lead to impressive performance on a wide range of tasks. At generation, these models are used auto-regressively, requiring a forward pass for each generated token, and thus reading the full set of parameters from memory. This is the bottleneck for generation, and it is often as fast to perform a forward pass on a couple of tokens as on a single one.
% \aj{great begining of abstract, the previous sentence contains 2 important information that are key to this project. I would rephrase it a bit to insist a bit more on them, and maybe even split it in 2 different sentences to make sure tha tboth messages is well received by the reader.}
% This lead to the development of speculative sampling, where a smaller model is used to draft a few tokens, that are then validated or rejected using a single forward pass of the large model. \aj{the use of a small model is not implied by the sentence preceeding the definition of speculative sampling} Unfortunately, this method requires two models that shares the same tokenizer\aj{not clear why it needs the same tokenizer from what is written above}, and thus limits its adoption. Here, we propose to use the same large model as the drafter, by using parallel decoding to generate candidates. We show promising performance, while requiring only a small number of additional parameters.\aj{I would give some specifics about model size and speedup}
% \end{abstract}

\section{Introduction}
% \eg{Add a few references in the intro. Also discuss why a model call for each generated token is problematic. I would not use bullet points here (and rephrase a bit to make it into a paragraph).}
% Over the recent few years,\aj{would drop the "Over [..] years} 
Since the Transformer architecture was proposed (\cite{vaswani2023attention}), large language models have achieved impressive results across natural language processing benchmarks~(\cite{brown2020language}). However, these remarkable achievements were only made possible because of dramatic increases in the number of parameters or model sizes (\cite{brown2020language}, \cite{wei2022emergent}), resulting in considerable memory requirements and greater processing times. 
% \aj{great 1st paragraph. I would rearrange it a bit to avoid digression (e.g., "while also influenced", or at least placing them after the core argument is made.}
This problem is further exacerbated by the fact that, at inference time, transformers are used auto-regressively: a new model call is needed for each generated token. This is especially problematic due to the memory-bandwidth cost of recurrently loading the model parameters and the past keys and values tensors (\cite{shazeer2019fast}).
% \aj{great paragraph that clearly define the problematic of this work. I would merge it with the first paragraph after trimming it from digression.}

Recently, several works~(\cite{chen2023accelerating,leviathan2023fast}) have proposed to reduce inference time 
by leveraging a smaller model to approximate generation from a larger model at a faster pace. 
The small model produces a few potential tokens, and the larger model evaluates all of the tokens at once in a single forward step. 
Importantly, the generation quality of the original large model is guaranteed by the rejection scheme that keeps only tokens that are generated with an identical distribution than the large model (\cite{chen2023accelerating}, \cite{leviathan2023fast}). 
While effective in practice, this approach requires to deploy simultaneously two models that share the same vocabulary, creating memory and running time bottlenecks.

An alternative solution is to directly leverage the large model to generate multiple tokens at once, instead of generating them auto-regressively.
This solution, called parallel decoding, can be implemented as a masked language model (\cite{ghazvininejad-etal-2019-mask}) or by copying the encoder input in the decoder in the context of encoder-decoder architectures (\cite{gu2018nonautoregressive}). 
These solutions have the advantage over speculative sampling of avoiding the need for a second model, but they require substantial changes to the Transformer architecture that make them not suitable as such for accelerating the decoding of a given pre-trained language model.

% \gm{As an alternative to the previous paragraph, I would propose the following: An alternative solution is to directly leverage the large model to generate multiple tokens in parallel, instead of generating them auto-regressively. This different strategy, called parallel decoding, does not require a second model and has been actualized as masked parallel decoding (\cite{ghazvininejad-etal-2019-mask}) or by copying the encoder input in the decoder in the context of encoder-decoder architectures (\cite{gu2018nonautoregressive}). However, approaches in this direction usually entail a loss in quality compared to auto-regressive models and require substantial changes to the Transformer architecture or to the way transformers are used.}

In this work, we propose to combine the best of both directions in a variant of the speculative sampling that we call Parallel Speculative Sampling (PaSS). 
The idea is to generate candidate tokens via parallel decoding by adding a small number of ``look-ahead embeddings'' and generate output for each of these additional embeddings. 
This solution does not require a second model, nor modifications to the large language model.
By design, our approach also generates at each step at least one token auto-regressively, guaranteeing the same loss-less quality of generations as speculating sampling methods.
The memory overhead of adding the extra embeddings is $O(d_{emb})$ new weights, that need to be trained.
This is several orders of magnitude smaller than any small model added by existing speculative sampling solutions. 
The most similar works to ours are \cite{stern2018blockwise} and \cite{medusa} where they add look-ahead classification heads instead of embeddings, leading to a worse memory overhead of $O(d_{emb}K)$ where $K$ is the vocabulary size.
Additionally, \cite{stern2018blockwise} focus solely on greedy decoding, and \cite{medusa} do not guarantee a loss-less decoding. 
Similarly, \cite{zhang2023draft} also drop the second model, but still decode auto-regressively.

\section{Method}
% \eg{Do not use capital letters for name of methods (except in titles).}

First, in section \ref{sps_subsection}, we briefly review the existing speculative sampling algorithm, as introduced by \cite{chen2023accelerating} and \cite{leviathan2023fast}. We then introduce our approach in section \ref{pass_subsection}.

\subsection{Speculative Sampling}\label{sps_subsection}
% \eg{I think the core idea is that it is significantly faster to compute a single forward pass on n tokens in parallel, than n forward passes sequentially.}

The goal of speculative sampling is to speed up the inference time of a target LLM. The core idea behind this algorithm is that it is significantly faster to compute a single forward pass on $n$ tokens in parallel than $n$ forward passes sequentially. To fulfil this objective, a second smaller and faster model, \emph{the drafter}, is used to generate a candidate sequence of tokens.
The length of the sequence is a hyper-parameter of the algorithm. 
The target LLM is then presented with all of the candidate tokens at once, in a single pass. 
The rejection scheme proposed by \cite{chen2023accelerating} and \cite{leviathan2023fast} guarantees that the distribution of the drafted tokens is the same as if they were generated in the first place by the target LLM. 
Additionally, due to the rejection scheme, one more token can be sampled after the sequence of accepted tokens from the logits gathered during the iteration model call. 
This ensures that, even if all draft tokens were rejected, the model call would still be of use. 
The steps are presented in detail in algorithm \ref{sps_algo} of the Appendix. 

% \aj{this sentence suggests that we have already introduced the existence of a larger model. I would spend a bit more time describing the problem we are trying to solve before jumping to the solutions}
\subsection{Parallel Speculative Sampling}\label{pass_subsection}

We propose a modified version of speculative sampling based on parallel decoding, that does not require a second model. 
The steps of our method are detailed in algorithm \ref{pass_algo}, but, importantly, each iteration of our algorithm requires two calls of the LLM:
\begin{itemize}
    \item \textbf{Drafting phase}: we call the model once to produce multiple tokens simultaneously using parallel decoding through look-ahead embeddings (see Sec.~\ref{drafting_subsection}). 
    The first generated token is not part of the draft to match the distribution in case of rejection.
    \item \textbf{Validation phase}: we call the model a second time to validate the draft (see Sec.~\ref{sps_subsection}).
    We sample a new token at the end of the sequence of accepted tokens, with no new model call.
\end{itemize}

The key behind our algorithm is that every call to the LLM adds at least one token to the final sequence of generated tokens.
This guarantees that the algorithm is at least as fast as generating from the LLM directly, and it also guarantees that we produce a correct sequence of tokens even in the case where additional tokens are rejected. 
On top of this, our algorithm can produce and accept at each iteration, multiple additional tokens, leading to a guaranteed speed up. 
Overall, the standard auto-regressive wall time of the target LLM is a lower bound to our method, while the upper bound is a speed-up of $(L + 2)/2 \times$, where $L$ is the number of tokens generated in the drafting phase. 
Our approach leverages the fact that the time required to process a single token or a small sequence of tokens does not differ significantly. This is because auto-regressive generation is mostly memory-bound, and processing additional tokens is thus negligible.

\begin{algorithm}
\small
    \caption{Parallel Speculative Sampling (PaSS) with Parallel Look-ahead Embeddings} \label{pass_algo}
    \begin{algorithmic}
        \STATE  Given $L$ look-ahead tokens $\texttt{[LA]}_1, \dots, \texttt{[LA]}_L$ and minimum target sequence length $T$. 
        \STATE  Given auto-regressive target model $q(.|.)$ and initial prompt sequence $x_{0}, \dots, x_{t}$.
        \STATE  Initialise $n \gets t$.
        \WHILE{$n < T$}
            \STATE In parallel, sample the next token $x_{n+1}$ and $L$ draft tokens $\tilde{x}_{1}$, $\dots$, $\tilde{x}_{L}$:
            \STATE{}
            \STATE{} \begin{center} \( x_{n+1} \sim q(x | x_1, \dots, x_n),\textit{ } \tilde{x}_{1} \sim q(x | x_1, \dots, x_n, \texttt{[LA]}_{1}),\textit{ }\dots,\textit{ } \tilde{x}_{L} \sim q(x | x_1, \dots, x_n, \texttt{[LA]}_{1}, \dots, \texttt{[LA]}_{L}) \) \end{center}
            \STATE{}
            \STATE Set $n \gets n+1$
            \STATE In parallel, compute $L + 1$ sets of logits from drafts $\tilde{x}_{1}$, $\dots$, $\tilde{x}_{L}$:
            \STATE{}
            \STATE{} \begin{center} \( q(x | x_1, \dots, x_{n}),\textit{ }q(x | x_1, \dots, x_{n}, \tilde{x}_{1}),\textit{ }\dots,\textit{ }q(x | x_1, \dots, x_{n}, \tilde{x}_{1}, \dots, \tilde{x}_{L}) \) \end{center}
            \STATE{}
            \FOR{$t = 1 : L$}
                \STATE  Sample $r \sim U[0, 1]$ from a uniform distribution.
                \vspace{0.5em}
                \IF{$\displaystyle{r < \min\left(1, \frac{q(\tilde{x}_t | x_1, \dots, x_{n-1}, \dots, x_{n+t-1})}{q(\tilde{x}_t | x_1, \dots, x_{n-1}, \texttt{[LA]}_{1}, \dots, \texttt{[LA]}_{t})}\right)}$}
                \vspace{0.5em}
                    \STATE  Set $x_{n+t} \gets \tilde{x}_t$ and $n \gets n + 1$
                \ELSE
                    \STATE  Sample
                    \STATE{}
                    \STATE{} \begin{center} \( x_{n+t} \sim (q(x | x_1, \dots, x_{n-1}, \dots, x_{n+t-1}) - q(x | x_1, \dots, x_{n-1}, \texttt{[LA]}_{1}, \dots, \texttt{[LA]}_{t}))_+ \) \end{center}
                    \STATE{}
                    \STATE and Exit for loop.
                \ENDIF
            \ENDFOR
            \STATE  If all $L$ tokens $x_{n+1}, \dots, x_{n+L}$ are accepted, sample extra token $x_{n+L+1} \sim q(x | x_1, \dots, x_{n+L})$ and set $n \gets n+1$.    
            
        \ENDWHILE
    \end{algorithmic}
\end{algorithm}

\subsubsection{Look-ahead embeddings} \label{drafting_subsection}

The target LLM is not trained to predict multiple tokens at once, and expect as input, the previously generated token.
In order to build this ability in the target LLM, we introduce  ``look ahead'' tokens, \texttt{[LA]}$_i$ for each look ahead position $i$ for 1 to $L$. 
A sequence of these tokens is added at the end of the input sequence and defines the number of steps that the model will predict ahead. 
In other words, we replace the original input sequence of tokens $(w_1,\dots,w_T)$ by a sequence with $L$ additional tokens, that is $(w_1,\dots,w_T,\texttt{[LA]}_1,\dots,\texttt{[LA]}_L)$.
This sequence is then processed with a single forward pass of the target model and produces for each new position a token from the original dictionary, i.e., without the extra \texttt{[LA]}$_i$ tokens.
This approach only requires learning the embeddings associated with the new tokens on a small training dataset and has a memory overhead of $Ld_{emb}$ parameters.

% \aj{is it sharing similarities with the CPC approach ( https://arxiv.org/pdf/1807.03748.pdf )? If yes, could be interesting to see how they present it} \gm{I'd say their approach is quite different}

\section{Experiments}

We test our method on two tasks: text and code completion. For each task, we use different non-overlapping datasets for the training of the look-ahead embeddings and the evaluation of our approach
(Sec.~\ref{data_subsection}). 
We also briefly describe baselines in Sec.~\ref{baselines_subsection} and report main results in Sec.~\ref{results_subsection}. 
All the experiments are run with a re-implementation of the 7B LLaMa model~(\cite{touvron2023llama}).

\subsection{Data} \label{data_subsection}

We use the 2023/02/20 English Wikipedia dump (\cite{wikidump}) for text completion and the Python split of The Stack corpus (\cite{Kocetkov2022TheStack}) for code completion.
We divide each dataset into training and test split. 
For the evaluation, we randomly sample 200 examples from the test split and use the 32 first tokens as prompts.
The maximum length for the generation is fixed at 512 tokens for all our experiments.
We use \textsc{top-k sampling}, with $k=10$ and a temperature of $0.8$ unless said otherwise.
For code completion, we also evaluate on the HumanEval benchmark~\citep{chen2021evaluating}, to validate that, as expected, our algorithm does not degrade the quality of generation.

\subsection{Baselines} \label{baselines_subsection}
We compare our method with two baselines.
\textbf{Autoregressive generation:} this baseline consists of autoregressively generating tokens from the LLM. We sample one token at a time using the KV cache for speedup.
\textbf{\texttt{[UNK]} as look-ahead token:} we apply our method with a fixed \texttt{[UNK]} token instead of trained look-ahead embeddings. We use the KV cache and update it after every model call according to the number of drafted tokens and the number of accepted tokens.

\subsection{Results} \label{results_subsection}

\noindent\textbf{Impact for different sampling schemes.} On the left panel of Table~\ref{tab:timing}, we compare the running time of our approach with the two baselines for different sampling schemes. 
We vary the temperature of the sampler from high variance in the generations (high temperature) or low variance (low temperature).
As expected, the speed-up is more important for lower temperatures, where the distribution of tokens is more peaky and easier to predict with an approximated scheme like PaSS.
We observe almost no gain compared to auto-regressive generation when using \texttt{[UNK]}.
Compared to the speed-up provided by PaSS, this shows that the finetuning of the look-ahead embeddings captures important information to predict future tokens.

\noindent\textbf{Impact of the number of look ahead embeddings.} On the right panel of Table~\ref{tab:timing}, we measure the impact of the number of look-head steps on our approach.
Running time decreases steadily up to 6 look-ahead steps, but more look-ahead steps annihilate the benefits of this approach.

\noindent\textbf{Impact of PaSS on final performance.} Finally, in Table~\ref{tab:perf}, we confirm that our decoding does not impact the performance of the model on 2 different generating tasks.
We only observe changes in performance that are below the margin of error, while improving the running time by up to 30\%.

\begin{table}
\caption{Time for generating a sequence of length 512 tokens, given a prompt of 32 tokens, as a function of temperature (left) and number of look-ahead tokens (right). We use 4 look-ahead tokens unless said otherwise. The results reported in the right table are on The Stack data.}
\centering
\begin{tabular}{lcccccc}
\toprule           & \multicolumn{3}{c}{The Stack} & \multicolumn{3}{c}{Wikipedia} \\
%\midrule
\cmidrule(lr){2-4} \cmidrule(lr){5-7}
Temperature        & 0.8 & 0.5 & 0.2 & 0.8 & 0.5 & 0.2 \\
\midrule
Auto-regressive            & 12.52 & 12.69 & 12.72 & 12.45 & 12.30 & 12.55 \\
\texttt{[UNK]} look-ahead  & 12.25 & 12.43 & 12.26 & 12.30 & 12.16 & 11.88 \\
PaSS                       & 9.79 & 9.46 & 8.96 & 10.23 & 9.78 & 9.43 \\
\bottomrule
\end{tabular} \hfill
\begin{tabular}{cc}
\toprule
\# LA tokens & Time \\
\midrule
2 & 10.03 \\
4 &  9.79 \\
6 &  9.66 \\
8 &  9.94 \\
\bottomrule
\end{tabular}
\label{tab:timing}
\end{table}

\begin{table}[h!]
\centering
\caption{Average time for generating one completion on the HumanEval dataset, as well as the \textsc{pass@n} metric. Following previous work, we use a temperature of 0.1 for \textsc{pass@1} and a temperature of 0.8 for \textsc{pass@10}. We use $k=25$ for \textsc{pass@10}. For PaSS, we use 4 look-ahead tokens.}
\begin{tabular}{lcccc}
\toprule
& \multicolumn{2}{c}{\textsc{pass@1}} & \multicolumn{2}{c}{\textsc{pass@10}} \\
& Time & Perf. & Time & Perf. \\
\midrule
Auto-regressive  & 10.52 sec & 13.2 \% & 10.15 sec & 22.5 \% \\
PaSS             & \phantom{0}7.17 sec & 13.4 \% & 8.17 sec & 22.5 \% \\
\bottomrule
\end{tabular}
\label{tab:perf}
\end{table}

%\begin{table}
%  \caption{Sample table title}
%  \label{results_table}
%  \centering
%  \begin{tabular}{lll}
%    \toprule
%    \multicolumn{2}{c}{Part}                   \\
%    \cmidrule(r){1-2}
%    Name     & Description     & Size ($\mu$m) \\
%    \midrule
%    Dendrite & Input terminal  & $\sim$100     \\
%    Axon     & Output terminal & $\sim$10      \\
%    Soma     & Cell body       & up to $10^6$  \\
%    \bottomrule
%  \end{tabular}
%\end{table}

% We summarize the key findings of these experiments below:

% \noindent\textbf{Increasing the number of look-ahead embeddings improves speed-up.} The best performing variant of our method is the one with the highest number of look-ahead embeddings. This confirms that processing a few more tokens does not significantly increase the model processing time, while, in some cases, leading to many tokens being accepted at once, which increases the speedup.   

\section{Conclusion}

We presented the parallel speculative sampling (PaSS) algorithm, a variant of the speculative sampling algorithm that does not require a second draft model: tokens are drafted in parallel through the use of masked-decoding via fine-tuned look-ahead embeddings. We showed that our method achieves significant speed-ups (up to $30\%$) by only learning as little as $O(d_{emb})$ additional weights. In future work, we want to explore how to improve the quality of parallel generation with look-ahead tokens, as we believe this is the most promising direction to improve performance of the PaSS algorithm.

\bibliographystyle{plainnat}
\bibliography{references}

\newpage
\section{Supplementary Material}

\subsection{Speculative Sampling Algorithm}
\begin{algorithm}
    \caption{Speculative Sampling (SpS) with Auto-Regressive Target and Draft Models} \label{sps_algo}
    \begin{algorithmic}
        \STATE  Given look-ahead $L$ and minimum target sequence length $T$. 
        \STATE  Given auto-regressive target model $q(.|.)$, auto-regressive draft model $p(.|.)$ and initial prompt sequence $x_{0}, \dots, x_{t}$.
        \STATE  Initialise $n \gets t$.
        \WHILE{$n < T$}
            \FOR{$t = 1 : L$}
                \STATE  Sample draft auto-regressively $\tilde{x}_{t} \sim p(x | x_1, \dots, x_n, \tilde{x}_{1}, \dots, \tilde{x}_{t-1})$
            \ENDFOR
            \STATE In parallel, compute $L + 1$ sets of logits from drafts $\tilde{x}_{1}$, $\dots$, $\tilde{x}_{L}$:
            \STATE{} \begin{center} \( q(x | x_1, \dots, x_n),\textit{ }q(x | x_1, \dots, x_n, \tilde{x}_{1}),\textit{ }\dots,\textit{ }q(x | x_1, \dots, x_n, \tilde{x}_{1}, \dots, \tilde{x}_{L}) \) \end{center}
            \FOR{$t = 1 : L$}
                \STATE  Sample $r \sim U[0, 1]$ from a uniform distribution.
                \IF{$r < \min\left(1, \frac{q(\tilde{x}_t | x_1, \dots, x_{n+t-1})}{p(\tilde{x}_t | x_1, \dots, x_{n+t-1})}\right)$}
                    \STATE  Set $x_{n+t} \gets \tilde{x}_t$ and $n \gets n + 1$
                \ELSE
                    \STATE  Sample $x_{n+t} \sim (q(x | x_1, \dots, x_{n+t-1}) - p(x | x_1, \dots, x_{n+t-1}))_+$ and exit for loop.
                \ENDIF
            \ENDFOR
            \STATE  If all tokens $x_{n+1}, \dots, x_{n+L}$ are accepted, sample extra token $x_{n+L+1} \sim q(x | x_1, \dots, x_{n+L})$ and set $n \gets n+1$.    
            
        \ENDWHILE
    \end{algorithmic}
\end{algorithm}

\subsection{Training details} \label{training_subsection}
%\begin{itemize}
%    \item Add training hyper parameters and details, as well as model details (?)
%\end{itemize}

For all our trainings (and evaluations), we load models in bfloat16. We freeze all the models parameters except for the new embeddings. For each batch, we randomly select a position where to insert the look-ahead embeddings and compute the loss only on the corresponding outputs. Before training, we initialize the new embeddings with the \textit{UNK} token embedding. We use the AdamW optimizer, with a learning rate of 0.01 and a batch size of 8 sequences. We train for 10k additional steps, with 2k warmup steps and a cosine learning rate schedule.

% Authors may wish to optionally include extra information (complete proofs, additional experiments and plots) in the appendix. All such materials should be part of the supplemental material (submitted separately) and should NOT be included in the main submission.

% \section*{References}

% References follow the acknowledgments in the camera-ready paper. Use unnumbered first-level heading for
% the references. Any choice of citation style is acceptable as long as you are
% consistent. It is permissible to reduce the font size to \verb+small+ (9 point)
% when listing the references.
% Note that the Reference section does not count towards the page limit.
% \medskip

% {
% \small

% [1] Alexander, J.A.\ \& Mozer, M.C.\ (1995) Template-based algorithms for
% connectionist rule extraction. In G.\ Tesauro, D.S.\ Touretzky and T.K.\ Leen
% (eds.), {\it Advances in Neural Information Processing Systems 7},
% pp.\ 609--616. Cambridge, MA: MIT Press.

% [2] Bower, J.M.\ \& Beeman, D.\ (1995) {\it The Book of GENESIS: Exploring
%   Realistic Neural Models with the GEneral NEural SImulation System.}  New York:
% TELOS/Springer--Verlag.

% [3] Hasselmo, M.E., Schnell, E.\ \& Barkai, E.\ (1995) Dynamics of learning and
% recall at excitatory recurrent synapses and cholinergic modulation in rat
% hippocampal region CA3. {\it Journal of Neuroscience} {\bf 15}(7):5249-5262.
% }

%%%%%%%%%%%%%%%%%%%%%%%%%%%%%%%%%%%%%%%%%%%%%%%%%%%%%%%%%%%%

\end{document}